\pdfoutput=1

\documentclass[11pt]{article}

\usepackage{authblk}
\usepackage{acl}

\usepackage{times}
\usepackage{latexsym}
\usepackage{siunitx}
\usepackage[T1]{fontenc}

\usepackage[utf8]{inputenc}

\usepackage{microtype}
\usepackage{todonotes}
\usepackage{booktabs}
\usepackage{listings}
\usepackage{hyperref}
\usepackage{graphicx}
\usepackage{tabularx}
\usepackage{multirow}
\usepackage{float}
\usepackage[symbol]{footmisc}
\title{Entities, Dates, and Languages: Zero-Shot on Historical Texts with T0} 

\author{\textbf{Francesco De Toni$^{1}$}, \textbf{Christopher Akiki$^{2}$},           \textbf{Javier de la Rosa$^{3}$}\\
    \textbf{Cl{\'e}mentine Fourrier$^{4}$}, \textbf{Enrique Manjavacas$^{5}$}, \textbf{Stefan Schweter$^{6}$}, \textbf{Daniel van Strien$^{7}$} \\
    \normalsize $^{1}$The University of Western Australia, \normalsize $^{2}$Leipzig University, \normalsize $^{3}$National Library of Norway, \normalsize $^{4}$Inria Paris\\
	\normalsize $^{5}$Leiden University, \normalsize $^{6}$Bayerische Staatsbibliothek, \normalsize $^{7}$British Library\\
	\normalsize{\texttt{francesco.detoni@uwa.edu.au}}
}

\begin{document}
\maketitle
\begin{abstract} 
\renewcommand{\thefootnote}{\fnsymbol{footnote}}
In this work, we explore whether the recently demonstrated zero-shot abilities of the T0 model extend to Named Entity Recognition for out-of-distribution languages and time periods. Using a historical newspaper corpus in 3 languages as test-bed, we use prompts to extract possible named entities. Our results show that a naive approach for prompt-based zero-shot multilingual Named Entity Recognition is error-prone, but highlights the potential of such an approach for historical languages lacking labeled datasets. Moreover, we also find that T0-like models can be probed to predict the publication date and language of a document, which could be very relevant for the study of historical texts\footnote{Authorship attribution (alphabetical): \S1: Akiki, De Toni, van Strien; \S2.1: Fourrier; \S2.2: Manjavacas; \S2.3 and experiment execution: Fourrier, de la Rosa, De Toni, Schweter; \S3: De Toni, Manjavacas; \S4: Akiki, van Strien; \S5: all the authors; Impacts Statement: Akiki, Fourrier, de la Rosa.}.
\end{abstract}

\setcounter{footnote}{0} 
\renewcommand{\thefootnote}{\arabic{footnote}}
\section{Introduction} 
This paper lies at the focal point of three orthogonal advances. First, the recent surge in GLAM\footnote{Galleries, libraries, archives, and museums.}-led digitisation efforts \cite{Terras2011}, open citizen science \cite{muki:2021citsci} and the expansive commodification of data \cite{hey:2003deluge}, have enabled a new mode of historical inquiry that capitalises on the `big data of the past' \cite{Kapland17}. Second, the 2017~breakthrough that was the transformer architecture \cite{VaswaniSPUJGKP17} has led to the so-called ImageNet moment of Natural Language Processing \cite{ruder2018nlpimagenet} and brought about unprecedented progress in transfer-learning \cite{raffel2020t5}, few-shot learning \cite{schick-schutze-2021-just}, zero-shot learning \cite{sanh2021multitask}, and prompt-based learning \cite{le-scao-rush-2021-many} for natural language. Third, the growing popularity of prompt-based methods \cite{LiuPengfei2021PPaP} has resulted in a new paradigm for training and fine-tuning Large Language Models~(LLM) as well as novel applications in Named Entity Recognition~(NER)~\cite{QaNER}.

NER for historical texts has been the focus of a growing body of research, most recently surveyed by~\citet{Ehrmann2021NamedER}. Both NER and the related task of Entity Linking can enhance our ability to search and navigate digitised historical materials~\cite{Neudecker2014LargescaleRO, Kim2015FindingNI}. However, applying NER to historical texts poses a number of challenges, including those due to errors in Optical Character Recognition (OCR)~\cite{Ehrmann2021NamedER,Hamdi2019AnAO,Boros2020AlleviatingDE} and domain transfer~\cite{Blouin_Favre_Auguste_Henriot_2021}. To advance research in this area, an increasing number of datasets have been created to support the development and evaluation of NER approaches in historical text~\cite{Neudecker2016AnOC, ehrmann_extended_2020,ehrmann_maud_2022_6089968}

In this paper, we examine the zero-shot abilities of T0---a prompt-based LLM developed as part of the BigScience project for open research \cite{sanh2021multitask}---on the challenging task of historical NER\footnote{\scriptsize{\url{https://github.com/bigscience-workshop/historical_texts}}}. This endeavour had two main hurdles: (1)~the model was neither trained to recognize entities, nor was it ever tested on that task; (2)~our evaluation dataset was out-of-distribution, containing both multilingual and historical data. 
To better contextualize the results of our experiments, we also run zero-shot prompt-based probing \cite{zhong-etal-2021-factual} to assess T0's broader ability of extracting factual knowledge about two key factors in our experiment, that is, language variation and historical variation in the dataset.

\begin{table*}[t]
    \centering
    \footnotesize
    \begin{tabular}{@{}c S[table-format=2.0]S[table-format=5.0]S[table-format=1.1]S[table-format=3.0]S[table-format=4.0]S[table-format=1.1]S[table-format=2.0]S[table-format=6.0]S[table-format=1.1]@{}} \toprule
         & \multicolumn{3}{c}{English} &
         \multicolumn{3}{c}{German} &
          \multicolumn{3}{c}{French} \\ 
        \cmidrule(r){2-4}\cmidrule(lr){5-7}\cmidrule(l){8-10}
        {Time period} & {\#Documents} & {\#Tokens} & {NE\%} 
        & {\#Documents} & {\#Tokens} & {NE\%} 
        & {\#Documents} & {\#Tokens} & {NE\%} \\ \midrule
        1790-1810 & 10 & 4143 & 3.1 & 13 & 6735 & 4.6 & 14 & 8550 & 4.4 \\
        1810-1830 & 15 & 4697 & 3.4 & 13 & 8049 & 2.6 & 10 & 12440 & 5.0 \\
        1830-1850 & 9 & 3974 & 4.0 & 19 & 15601 & 2.8 & 10 & 11659 & 3.9 \\
        1850-1870 & 0 & 0 & {-} & 21 & 16021 & 3.8 & 9 & 10321 & 3.9 \\
        1870-1890 & 7 & 2202 & 1.9 & 16 & 17181 & 3.7 & 15 & 16272 & 4.2 \\
        1890-1910 & 12 & 4509 & 2.9 & 12 & 12829 & 4.3 & 19 & 16874 & 4.6 \\
        1910-1930 & 13 & 5499 & 3.1 & 13 & 18134 & 3.3 & 30 & 30403 & 3.8 \\
        1930-1950 & 3 & 520 & 4.2 & 29 & 24566 & 5.7 & 32 & 35962 & 4.2 \\
        \midrule
        Total & 69 & 25544 & 3.2 & 136 & 119116 & 4.0 & 139 & 142481 & 4.2 \\
        \bottomrule
    \end{tabular}
    \caption{Data description: splits by date and language of the CLEF-HIPE 2020 dataset.} 
    \label{tab:data_desc}
\end{table*}

\section{Experimental setup}
\subsection{Data description}
\label{sec:Data description}
Our data comes from version~1.4 of the CLEF-HIPE\footnote{Conference and Labs of the Evaluation Forum - \\ Identifying Historical People, Places and other Entities.} 2020 open-access dataset\footnote{\scriptsize{\url{https://github.com/impresso/CLEF-HIPE-2020}}}: an OCR'ed newspaper corpus annotated for NER \cite{ehrmann_extended_2020}. It contains Swiss and Luxembourgish newspapers from 1790 to 2010, in English, German and French. For our experiment, we use only entities of coarse type, according to their literal sense. Coarse entity types in the CLEF-HIPE 2020 dataset are persons, locations, organizations, dates and products (which includes media and doctrines). 

We mix the original training and validation sets to constitute our test set\footnote{For English, we use only the validation set, as the training set is absent}, and we split this new set by language and date (using 20~years time intervals,\footnote{We chose 20-year spans as the smallest time range producing somewhat balanced splits.} see Table~\ref{tab:data_desc}). Each language dataset is relatively balanced between 1810 and 1910, with English containing between 2,202 and 4,697 tokens per split with the exception of one split (1850-1870  English) for which there are no tokens. German contains between 6,735 and 12,829~tokens, and French contains between 8,550 and 16,874~tokens. The end periods contain on average more tokens for German and French. Overall, the dataset contains 3.8\% of named entities (from 1.9 to 5.6\%, depending on time periods and datasets). The most balanced dataset across time periods is the French one (between 3.8 and 4.6\% named entities). 

\subsection{Model description}
In our experiments, we use the T0++ variant of the T0 language model \cite{sanh2021multitask}, based on the LM-adapted T5 model \cite{lester-etal-2021-power}, itself a variant of the T5 model \cite{raffel2020t5}, which further pretrains the original encoder-decoder architecture of T5 with an autoregressive language modeling objective.\footnote{The added specific pretraining of T0 uses a set of 11 varied tasks represented by a total of 55 datasets.} Crucially, this pretraining is done using a prompt-based training setup, in which training examples are transformed into prompts using a variety of crowd-sourced prompt templates. This setup allows T0 to perform few-shot and zero-shot learning when presented with new prompts for a previously unseen task.

\subsection{Experiments}
\label{sec:Experimental setup}
Our goal in this paper is to see if and how state-of-the-art language models can be used for historical NLP tasks, with minimal modifications and fine-tuning.\footnote{\label{ft:ecology}Ecological concerns and funding inequalities raise considerations on how to best use already existing models for lower-resourced tasks, and with spending as little further computing power in fine-tuning as possible \cite{bender-etal-2021-stochastic}.} As such, we choose to use a `naive' approach, by directly asking the model which named entities a given sentence contains. To do so, we first design prompts for each named entity type (see Table~\ref{tab:example_table}). For each sentence in the dataset, we then 1) use all the generation prompts to determine if the sentence contains named entities of each entity type \footnote{For PROD entities, the generation prompt explicitly mentioned \textit{media} and \textit{doctrines}, as we regarded the word \textit{product} as too generic to return an accurate answer from T0.}; 2) filter the model's answer to keep only tokens that are actually in the input sentence, keeping the entity covering the longer span in case of nested entities; and 3) ask a disambiguation question if needed (if a token was assigned to multiple entities by the model). Results are stored at each step.

\begin{table*}[t]
    \centering
    \footnotesize
    \begin{tabular}{c|l} \toprule
        Entity & \multicolumn{1}{c}{Step (1) Generation prompt} \\  \midrule
        PERS & Input: \textbf{\texttt{<sentence>}}\textbf{\texttt{\textbackslash n}} In input, what are the names of person? Separate answers with commas. \\
        LOC & Input: \textbf{\texttt{<sentence>}}\textbf{\texttt{\textbackslash n}} In input, what are the names of location? Separate answers with commas. \\
        PROD & Input: \textbf{\texttt{<sentence>}}\textbf{\texttt{\textbackslash n}} In input, what are the names of media or doctrine? Separate answers with commas. \\ \midrule
        Entities & \multicolumn{1}{c}{Step (3) Disambiguation prompt} \\ \midrule
        PERS, LOC & Input: \textbf{\texttt{<sentence>}}\textbf{\texttt{\textbackslash n}} In input, is \textbf{\texttt{<entity>}} a person or a location? Give only one answer. \\ \midrule
        Fact & \multicolumn{1}{c}{Factual probing prompts} \\ \midrule
        Language & \textbf{\texttt{<sentence>\textbackslash n}} Q:Name the language of the previous sentence.\texttt{\textbf{\textbackslash n}}A: \\
        Date & In which year is the following text likely to have been published: text: \textbf{\texttt{<text>}} \\
        \bottomrule
    \end{tabular}
    \caption{Example prompts for generation and disambiguation (Sec.~\ref{sec:Experimental setup}), as well as factual probing (Sec.~\ref{sec:probing}).}
    \label{tab:example_table}
\end{table*}

We then evaluate the results and conduct two additional experiments to better understand the impact of the dataset language and time period on the performance of the LM.

\section{Results}
\subsection{Limitations}
Results reveal limitations in our proposed approach. First, T0 exhibits a clear tendency to produce non-empty outputs regardless of the presence or absence of named entities in the input: none of the prompts generates an empty answer. This is especially visible for the entity PROD, for which T0 answers over 55\% of the queries with the name of the entity itself (e.g. either \textit{media} or \textit{doctrine}) rather than with any other token from the input sentence. Second, adequately matching T0's output with tokens in the input sentence proved difficult. Even when T0 generates an answer semantically very close to the correct token in the sentence, differences in spelling prevent the algorithm from correctly associating T0's answer with said token in the input sentence. This problem is inherent to the nature of our dataset: frequent OCR errors generate unpredictable variations in `gold' word spelling (including spacing between words and letters or diacritics variation), which are automatically corrected by T0 during its predictions,\footnote{E.g.~Respelling words that were garbled due to noisy OCR.} which negatively affects our ability to automatically match its answers with corresponding tokens in the sentence. In other instances, the model translated words from French and German into English. Further experiments might need to mitigate language variety by adding input text to the prompt, to help the model correctly assess the language in which it must answer. As all answers predicted are considered strictly incorrect, the algorithm never enters its disambiguation phase. We therefore analyse non disambiguated results. 

\subsection{Evaluation}
\begin{figure*}[t]
    \includegraphics[width=\textwidth]{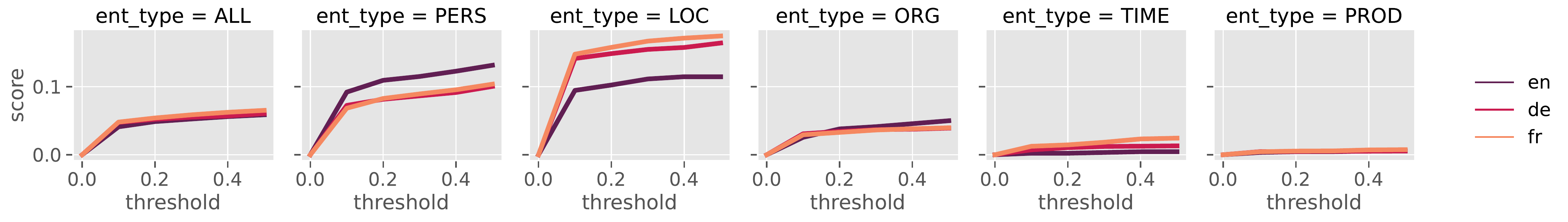}
    \caption{Precision for the different languages at different Levenshtein distance thresholds. Languages are distinguished by the line color.}
    \label{fig:Lev_thresholds_prec}
\end{figure*}
\begin{figure*}[ht]
    \includegraphics[width=\textwidth]{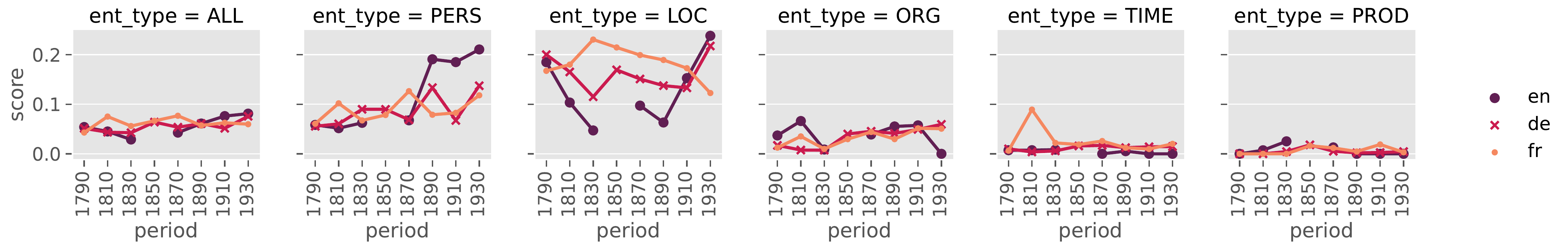}
    \caption{Precision for the different languages at Levenshtein threshold 0.4 across periods. \newline Languages are distinguished by both the line color and the type of dot.}
    \label{fig:Lev_threshold_04_period_prec}
\end{figure*}

To evaluate proximity between predictions and gold, we compare `gold' tokens with predicted tokens using normalized Levenshtein distance,\footnote{Normalization was done with regard to the length of the longest token (predicted or correct), and results were kept below a threshold. We tried 0.0, 0.1, 0.2, 0.3, 0.4 and 0.5.} using this metric as a proxy to identify best predictions for each entity query in each sentence. For a given example, we define (1) the true positive as the prediction with the shortest Levenshtein distance from the gold; (2) false positives as predictions of entities that are not actually present in the input sentence; and (3) false negatives as predictions that have longer Levenshtein distance to the gold tokens (i.e. predictions that would have failed to identify entity tokens in the sentence). Precision and F1-score are relatively low, especially for PROD entities, which were the most difficult to define in terms of text prompts. Higher values for recall are due to the fact that increasing the Levenshtein threshold makes it more likely to find an acceptable answer among those generated by T0. Unsurprisingly, the highest increase is found in TIME entities (dates have fixed formats, which makes it more likely to find an acceptable distance between predictions and correct tokens). Precision scores for each entity type are shown in Figure~\ref{fig:Lev_thresholds_prec} (see Fig.~\ref{fig:Lev_thresholds} in Appendix for recall and F1-score). The results of our experiment suggest that, although T0 struggles to return exact matches of the entities in the input sentence, it is still capable of generating answers that are semantically close to the correct tokens.

After manually inspecting the dataset and its numerous OCR artifacts, we choose 0.4 as a reasonable heuristic of close semantic similarity between T0's output and gold tokens. We find that using a threshold of 0.4 prevents the apparition of false positives, and therefore we use it to analyze differences between languages and between historical periods within the dataset. With respect to variations across languages, we observe that the precision of predictions in English does not have a clear edge over precision in French and German (Fig.~\ref{fig:Lev_threshold_04_period_prec}; see also Fig.~\ref{fig:Lev_threshold_04} in Appendix). This is unexpected, as T0 should display considerable bias towards English, which constitutes most of its training data. 
With respect to variations across periods, we observe an improvement in precision (and F1-score) for PERS and LOC entities in English texts from 1850s onwards (Fig.~\ref{fig:Lev_thresholds}; for recall and F1-score, see Fig.~\ref{fig:Lev_threshold_04_period} in Appendix), when for other entities and languages, precision and F1-score are either stable or show a downward trend (e.g. LOC in German)\footnote{The absence of documents in the 1850-1870 English split explains the missing values for English in that period.}. Variations in recall cannot be reduced to clear trends, but they are particularly erratic in English texts. A possible explanation could be that T0 is more sensitive to English text inputs, and therefore outputs a higher or lower number of irrelevant answers based on the specific content of each input sentence.

Baseline comparison with the results of the HIPE 2020 evaluation campaign\footnote{\scriptsize{\url{https://github.com/impresso/CLEF-HIPE-2020/blob/master/evaluation-results/ranking\_summary\_final.md}}} confirms that our implementation of zero-shot NER with T0 is below SOTA performance. As baselines, we considered the micro precision, recall and F1-score of coarse NER (literal sense) with fuzzy boundary matching from HIPE 2020 (see Table~\ref{tab:HIPE_results}).

\begin{table}[h!]
\centering
\begin{tabular}{l|rrr}
\toprule
Languages &    Precision &  Recall & F1-score \\
\midrule
English & 0.794 & 0.817 & 0.806\\
German & 0.870 & 0.886 & 0.878\\
French & 0.912 & 0.931 & 0.921\\
\bottomrule
\end{tabular}
\caption{HIPE 2020's best results for coarse NER (literal) with fuzzy boundary.}
\label{tab:HIPE_results}
\end{table}

All the scores from our experiments with T0 are below the best results from HIPE 2020. We note that the results from HIPE 2020 are based on experiments conducted on the HIPE test sets in each language (these are different from the test sets we used in our experiments, for which we combined the original HIPE training and validation sets; see Sec.~\ref{sec:Data description}). For this reason, we re-run our experiments on the original HIPE test sets, keeping the threshold for Levenshtein distance at 0.4. We observe no significant improvement in precision and F1-score compared to the results of our experiments on the combined training and validation sets. We observe some improvements in recall, especially for English and for TIME, with recall reaching 1.0 for some combinations of language, entity and time period. However, we believe that this improvement is not significant and it is due to our choice of the Levenshtein threshold, as already explained above.

\section{Prompt-based factual probing}
\label{sec:probing}
In addition to our main experiment on NER, we run two further experiments to assess T0's ability to do inference in a multilingual setting and to identify historical variation in textual corpora.
\paragraph{Probing for language}
To gauge T0's ability to reason in a multilingual setting, we test the model's language identification ability. To that end, we use a trilingual\footnote{French, German, and English; 1000 sentences each.} subset of the WiLI-2018 - Wikipedia Language Identification dataset \citep{thoma:2018-wili} and prompt the model on language (Table~\ref{tab:example_table}). 
We find that the model is able to correctly classify 83\% of French sentences, 74.1\% of German sentences, but only 35.4\% of English sentences. The previously mentioned potential sensitivity of the model to its own mother tongue might explain this result.

\paragraph{Probing for publication date}
To assess T0's treatment of historical text, we study how well it predicts the likely date of publication for a piece of text from our test dataset by prompting on publication date (Table~\ref{tab:example_table}). 

\begin{table}[h!]
\centering
\begin{tabular}{l|rr}
\toprule
& \multicolumn{2}{c}{Absolute errors} \\
Languages &    Mean &  Median  \\
\midrule
English &  40.48 &                   30.0 \\
German &  40.11 &                   32.0 \\
French &  55.25 &                   48.0 \\
\bottomrule
\end{tabular}
\caption{Date prediction results.}
\label{tab:Publication_prediction}
\end{table}

Table~\ref{tab:Publication_prediction} shows the prediction errors. Subtle language change can occur in a measurable way in as short a period as a decade \cite{Juola_2003}, and therefore a median absolute error of 30 suggests that T0 is good in predicting publication dates. We notice some variation in performance between different languages, with French performing slightly worse on both metrics (possibly because it belongs to a different language family from English, contrary to German).

\section{Conclusion}
We have presented our experiment to evaluate T0 for zero-shot historical NER, as well as on the prediction of language and publication date of historical texts. Our results show that historical texts present additional challenges for zero-shot NER (especially because historical datasets often include noisy OCR), but that T0 can however be used as is for language and date prediction. Next steps will be experimenting on different prompts and matching methods, as well as testing few-shot NER.

\section*{Acknowledgements}
This work took place under the umbrella of the ``Language Models for Historical Texts'' working group of the BigScience ``Summer of Language Models 21'' workshop\footnote{\url{https://bigscience.huggingface.co/}}. We are thankful to the organizers of this workshop for providing a forum conducive to collaborative and open scientific inquiry. We are especially grateful to Suzana Ili{\'c} for her help setting up and organising the working group.

\section*{Broader Impacts Statement}
In this paper, we take exploratory first steps toward instrumentalising the T0 large language model on the task of historical NER. We deem it appropriate to briefly discuss the ethical considerations that are implied by such a usage. First, if a model can be used in a context for which it was not explicitly intended for, it stands to reason that it can be misused in that same context: while recognizing entities in historical texts might at first glance seem innocuous, numerous studies focused on BIPOC representation in history have shown that this is not the case, as some marginalized groups tend to suffer from history erasure \cite{kellow1999erasing,ram2020black,stanley2021beyond}. Second, the automation and scaling of historical inquiry could potentially lead to unreflected (mis)interpretations of the past \cite{gibbs2013hermeneutics,gibbs2016new}. Third, the experimental nature of prompt-based inference could lead to a considerable carbon footprint, owing to the trial-and-error nature of manual prompt calibration, though this cost would still be lower than training a new model from scratch or fine-tuning an existing LLM (see footnote~\ref{ft:ecology}).

\bibliography{anthology,custom}

\begin{thebibliography}{31}
\expandafter\ifx\csname natexlab\endcsname\relax\def\natexlab#1{#1}\fi

\bibitem[{Baptiste et~al.(2021)Baptiste, Favre, Auguste, and
  Henriot}]{Blouin_Favre_Auguste_Henriot_2021}
Blouin Baptiste, Benoit Favre, Jeremy Auguste, and Christian Henriot. 2021.
\newblock \href {https://hal.archives-ouvertes.fr/hal-03550384} {{Transferring
  Modern Named Entity Recognition to the Historical Domain: How to Take the
  Step?}}
\newblock In \emph{{Workshop on Natural Language Processing for Digital
  Humanities (NLP4DH)}}, Silchar (Online), India.

\bibitem[{Bender et~al.(2021)Bender, Gebru, McMillan-Major, and
  Shmitchell}]{bender-etal-2021-stochastic}
Emily~M. Bender, Timnit Gebru, Angelina McMillan-Major, and Shmargaret
  Shmitchell. 2021.
\newblock \href {https://doi.org/10.1145/3442188.3445922} {On the dangers of
  stochastic parrots: Can language models be too big?}
\newblock In \emph{Proceedings of the 2021 ACM Conference on Fairness,
  Accountability, and Transparency}, FAccT '21, pages 610--623, New York, NY,
  USA. Association for Computing Machinery.

\bibitem[{Boros et~al.(2020)Boros, Hamdi, Linhares~Pontes, Cabrera-Diego,
  Moreno, Sidere, and Doucet}]{Boros2020AlleviatingDE}
Emanuela Boros, Ahmed Hamdi, Elvys Linhares~Pontes, Luis~Adri{\'a}n
  Cabrera-Diego, Jose~G. Moreno, Nicolas Sidere, and Antoine Doucet. 2020.
\newblock \href {https://doi.org/10.18653/v1/2020.conll-1.35} {Alleviating
  digitization errors in named entity recognition for historical documents}.
\newblock In \emph{Proceedings of the 24th Conference on Computational Natural
  Language Learning}, pages 431--441, Online. Association for Computational
  Linguistics.

\bibitem[{Ehrmann et~al.(2021)Ehrmann, Hamdi, Pontes, Romanello, and
  Doucet}]{Ehrmann2021NamedER}
Maud Ehrmann, Ahmed Hamdi, Elvys~Linhares Pontes, Matteo Romanello, and Antoine
  Doucet. 2021.
\newblock \href {http://arxiv.org/abs/2109.11406} {Named entity recognition and
  classification on historical documents: {A} survey}.
\newblock \emph{CoRR}, abs/2109.11406.

\bibitem[{Ehrmann et~al.(2022)Ehrmann, Romanello, Doucet, and
  Clematide}]{ehrmann_maud_2022_6089968}
Maud Ehrmann, Matteo Romanello, Antoine Doucet, and Simon Clematide. 2022.
\newblock \href {https://doi.org/10.5281/zenodo.6089968} {{HIPE}-2022 shared
  task named entity datasets}.

\bibitem[{Ehrmann et~al.(2020)Ehrmann, Romanello, Fl{\"u}ckiger, and
  Clematide}]{ehrmann_extended_2020}
Maud Ehrmann, Matteo Romanello, Alex Fl{\"u}ckiger, and Simon Clematide. 2020.
\newblock \href {https://doi.org/10.5281/zenodo.4117566} {Extended {Overview}
  of {CLEF HIPE} 2020: Named entity processing on historical newspapers}.
\newblock In \emph{{CLEF 2020 Working Notes}. {Working Notes} of {CLEF} 2020 -
  {Conference} and {Labs} of the {Evaluation Forum}}, volume 2696, page~38,
  {Thessaloniki, Greece}. {CEUR-WS}.

\bibitem[{Gibbs and Owens(2013)}]{gibbs2013hermeneutics}
Fred Gibbs and Trevor Owens. 2013.
\newblock \href
  {https://doi.org/http://dx.doi.org/10.3998/dh.12230987.0001.001} {The
  hermeneutics of data and historical writing}.
\newblock In Kristen Nawrotzki and Jack Dougherty, editors, \emph{Writing
  History in the Digital Age}, pages 159--172. University of {Michigan}
  {Press}.

\bibitem[{Gibbs(2016)}]{gibbs2016new}
Frederick~W Gibbs. 2016.
\newblock New forms of history: Critiquing data and its representations.
\newblock \emph{The American Historian}, 7:31--36.

\bibitem[{Haklay et~al.(2021)Haklay, Fraisl, Tzovaras, Hecker, Gold, Hager,
  Ceccaroni, Kieslinger, Wehn, Woods, Nold, Balázs, Mazzonetto, Ruefenacht,
  Shanley, Wagenknecht, Motion, Sforzi, Riemenschneider, and
  Vohland}]{muki:2021citsci}
Muki Haklay, Dilek Fraisl, Bastian Tzovaras, Susanne Hecker, Margaret Gold,
  Gerid Hager, Luigi Ceccaroni, Barbara Kieslinger, Uta Wehn, Sasha Woods,
  Christian Nold, Bálint Balázs, Marzia Mazzonetto, Simone Ruefenacht, Lea
  Shanley, Katherin Wagenknecht, Alice Motion, Andrea Sforzi, Dorte
  Riemenschneider, and Katrin Vohland. 2021.
\newblock \href {https://doi.org/10.1098/rsos.202108} {Contours of citizen
  science: a vignette study}.
\newblock \emph{Royal Society Open Science}, 8:202108.

\bibitem[{Hamdi et~al.(2019)Hamdi, Jean-Caurant, Sid{\`e}re, Coustaty, and
  Doucet}]{Hamdi2019AnAO}
Ahmed Hamdi, Axel Jean-Caurant, Nicolas Sid{\`e}re, Micka{\"e}l Coustaty, and
  Antoine Doucet. 2019.
\newblock An analysis of the performance of named entity recognition over ocred
  documents.
\newblock \emph{2019 ACM/IEEE Joint Conference on Digital Libraries (JCDL)},
  pages 333--334.

\bibitem[{Hey and Trefethen(2003)}]{hey:2003deluge}
Tony Hey and Anne Trefethen. 2003.
\newblock \href {https://doi.org/https://doi.org/10.1002/0470867167.ch36}
  {\emph{The Data Deluge: An e-Science Perspective}}, chapter~36. John Wiley \&
  Sons, Ltd.

\bibitem[{Juola(2003)}]{Juola_2003}
Patrick Juola. 2003.
\newblock \href {https://doi.org/10.1023/A:1021839220474} {The time course of
  language change}.
\newblock \emph{Computers and the Humanities}, 37(1):77–96.

\bibitem[{Kaplan and Di~Lenardo(2017)}]{Kapland17}
Fr{\'{e}}d{\'{e}}ric Kaplan and Isabella Di~Lenardo. 2017.
\newblock \href {https://doi.org/10.3389/fdigh.2017.00012} {Big data of the
  past}.
\newblock \emph{Frontiers Digit. Humanit.}, 4:12.

\bibitem[{Kellow(1999)}]{kellow1999erasing}
Margaret~MR Kellow. 1999.
\newblock Erasing slavery: Memory, history, and race in new england.
\newblock \emph{Reviews in American History}, 27(4):526--533.

\bibitem[{Kim and Cassidy(2015)}]{Kim2015FindingNI}
Sunghwan~Mac Kim and Steve Cassidy. 2015.
\newblock \href {https://aclanthology.org/U15-1007} {Finding names in {T}rove:
  Named entity recognition for {A}ustralian historical newspapers}.
\newblock In \emph{Proceedings of the Australasian Language Technology
  Association Workshop 2015}, pages 57--65, Parramatta, Australia.

\bibitem[{Le~Scao and Rush(2021)}]{le-scao-rush-2021-many}
Teven Le~Scao and Alexander Rush. 2021.
\newblock \href {https://doi.org/10.18653/v1/2021.naacl-main.208} {How many
  data points is a prompt worth?}
\newblock In \emph{Proceedings of the 2021 Conference of the North American
  Chapter of the Association for Computational Linguistics: Human Language
  Technologies}, pages 2627--2636, Online. Association for Computational
  Linguistics.

\bibitem[{Lester et~al.(2021)Lester, Al-Rfou, and
  Constant}]{lester-etal-2021-power}
Brian Lester, Rami Al-Rfou, and Noah Constant. 2021.
\newblock \href {https://doi.org/10.18653/v1/2021.emnlp-main.243} {The power of
  scale for parameter-efficient prompt tuning}.
\newblock In \emph{Proceedings of the 2021 Conference on Empirical Methods in
  Natural Language Processing}, pages 3045--3059, Online and Punta Cana,
  Dominican Republic. Association for Computational Linguistics.

\bibitem[{Liu et~al.(2022)Liu, Xiao, Zhu, Zhang, Li, and Arnold}]{QaNER}
Andy~T. Liu, Wei Xiao, Henghui Zhu, Dejiao Zhang, Shang-Wen Li, and Andrew
  Arnold. 2022.
\newblock \href {http://arxiv.org/abs/arXiv:2203.01543} {Qa{NER}: Prompting
  question answering models for few-shot named entity recognition}.

\bibitem[{Liu et~al.(2021)Liu, Yuan, Fu, Jiang, Hayashi, and
  Neubig}]{LiuPengfei2021PPaP}
Pengfei Liu, Weizhe Yuan, Jinlan Fu, Zhengbao Jiang, Hiroaki Hayashi, and
  Graham Neubig. 2021.
\newblock \href {http://arxiv.org/abs/2107.13586} {Pre-train, prompt, and
  predict: {A} systematic survey of prompting methods in natural language
  processing}.

\bibitem[{Neudecker(2016)}]{Neudecker2016AnOC}
Clemens Neudecker. 2016.
\newblock \href {https://aclanthology.org/L16-1689} {An open corpus for named
  entity recognition in historic newspapers}.
\newblock In \emph{Proceedings of the Tenth International Conference on
  Language Resources and Evaluation ({LREC}'16)}, pages 4348--4352,
  Portoro{\v{z}}, Slovenia. European Language Resources Association (ELRA).

\bibitem[{Neudecker et~al.(2014)Neudecker, Wilms, Faber, and van
  Veen}]{Neudecker2014LargescaleRO}
Clemens Neudecker, Lotte Wilms, Willem~Jan Faber, and Theo van Veen. 2014.
\newblock \href
  {https://www.ifla.org/wp-content/uploads/2019/05/assets/newspapers/Geneva_2014/s6-neudecker_faber_wilms-en.pdf}
  {Large-scale refinement of digital historic newspapers with named entity
  recognition}.
\newblock In \emph{IFLA Congress 2014 – Digital Transformation and the
  Changing Role of News Media in the 21st Century}.

\bibitem[{Raffel et~al.(2020)Raffel, Shazeer, Roberts, Lee, Narang, Matena,
  Zhou, Li, and Liu}]{raffel2020t5}
Colin Raffel, Noam Shazeer, Adam Roberts, Katherine Lee, Sharan Narang, Michael
  Matena, Yanqi Zhou, Wei Li, and Peter~J. Liu. 2020.
\newblock \href {http://jmlr.org/papers/v21/20-074.html} {Exploring the limits
  of transfer learning with a unified text-to-text transformer}.
\newblock \emph{Journal of Machine Learning Research}, 21(140):1--67.

\bibitem[{Ram(2020)}]{ram2020black}
Christelle Ram. 2020.
\newblock \href {https://scholarship.rollins.edu/honors/121/} {\emph{Black
  historical erasure: A critical comparative analysis in {R}osewood and
  {O}coee}}.
\newblock Ph.D. thesis, Rollins College.

\bibitem[{Ruder(2018)}]{ruder2018nlpimagenet}
Sebastian Ruder. 2018.
\newblock {NLP's ImageNet moment has arrived}.
\newblock \url{https://ruder.io/nlp-imagenet/}.

\bibitem[{Sanh et~al.(2021)Sanh, Webson, Raffel, Bach, Sutawika, Alyafeai,
  Chaffin, Stiegler, Scao, Raja, Dey, Bari, Xu, Thakker, Sharma, Szczechla,
  Kim, Chhablani, Nayak, Datta, Chang, Jiang, Wang, Manica, Shen, Yong, Pandey,
  Bawden, Wang, Neeraj, Rozen, Sharma, Santilli, F{\'{e}}vry, Fries, Teehan,
  Biderman, Gao, Bers, Wolf, and Rush}]{sanh2021multitask}
Victor Sanh, Albert Webson, Colin Raffel, Stephen~H. Bach, Lintang Sutawika,
  Zaid Alyafeai, Antoine Chaffin, Arnaud Stiegler, Teven~Le Scao, Arun Raja,
  Manan Dey, M.~Saiful Bari, Canwen Xu, Urmish Thakker, Shanya Sharma, Eliza
  Szczechla, Taewoon Kim, Gunjan Chhablani, Nihal~V. Nayak, Debajyoti Datta,
  Jonathan Chang, Mike~Tian{-}Jian Jiang, Han Wang, Matteo Manica, Sheng Shen,
  Zheng~Xin Yong, Harshit Pandey, Rachel Bawden, Thomas Wang, Trishala Neeraj,
  Jos Rozen, Abheesht Sharma, Andrea Santilli, Thibault F{\'{e}}vry, Jason~Alan
  Fries, Ryan Teehan, Stella Biderman, Leo Gao, Tali Bers, Thomas Wolf, and
  Alexander~M. Rush. 2021.
\newblock \href {http://arxiv.org/abs/2110.08207} {Multitask prompted training
  enables zero-shot task generalization}.
\newblock \emph{CoRR}, abs/2110.08207.

\bibitem[{Schick and Sch{\"u}tze(2021)}]{schick-schutze-2021-just}
Timo Schick and Hinrich Sch{\"u}tze. 2021.
\newblock \href {https://doi.org/10.18653/v1/2021.naacl-main.185} {It{'}s not
  just size that matters: Small language models are also few-shot learners}.
\newblock In \emph{Proceedings of the 2021 Conference of the North American
  Chapter of the Association for Computational Linguistics: Human Language
  Technologies}, pages 2339--2352, Online. Association for Computational
  Linguistics.

\bibitem[{Stanley(2021)}]{stanley2021beyond}
Michelle~A Stanley. 2021.
\newblock \emph{Beyond erasure: Indigenous genocide denial and settler
  colonialism}.
\newblock Routledge.

\bibitem[{Terras(2011)}]{Terras2011}
Melissa~M. Terras. 2011.
\newblock \href {https://doi.org/10.1007/978-94-6091-299-3_1} {The rise of
  digitization}.
\newblock In Ruth Rikowski, editor, \emph{Digitisation Perspectives}, pages
  3--20. Sense Publishers, Rotterdam.

\bibitem[{Thoma(2018)}]{thoma:2018-wili}
Martin Thoma. 2018.
\newblock \href {https://doi.org/10.5281/zenodo.841984} {{WiLI-2018 - Wikipedia
  Language Identification database}}.

\bibitem[{Vaswani et~al.(2017)Vaswani, Shazeer, Parmar, Uszkoreit, Jones,
  Gomez, Kaiser, and Polosukhin}]{VaswaniSPUJGKP17}
Ashish Vaswani, Noam Shazeer, Niki Parmar, Jakob Uszkoreit, Llion Jones,
  Aidan~N. Gomez, Lukasz Kaiser, and Illia Polosukhin. 2017.
\newblock \href
  {https://proceedings.neurips.cc/paper/2017/hash/3f5ee243547dee91fbd053c1c4a845aa-Abstract.html}
  {Attention is all you need}.
\newblock In \emph{Advances in Neural Information Processing Systems 30: Annual
  Conference on Neural Information Processing Systems 2017, December 4-9, 2017,
  Long Beach, CA, {USA}}, pages 5998--6008.

\bibitem[{Zhong et~al.(2021)Zhong, Friedman, and
  Chen}]{zhong-etal-2021-factual}
Zexuan Zhong, Dan Friedman, and Danqi Chen. 2021.
\newblock \href {https://doi.org/10.18653/v1/2021.naacl-main.398} {Factual
  probing is [{MASK}]: Learning vs. learning to recall}.
\newblock In \emph{Proceedings of the 2021 Conference of the North American
  Chapter of the Association for Computational Linguistics: Human Language
  Technologies}, pages 5017--5033, Online. Association for Computational
  Linguistics.

\end{thebibliography}
\bibliographystyle{acl_natbib}

\newpage
\onecolumn
\appendix
\label{sec:appendix}

\section*{Appendix: Full scores of Levenshtein distance}
The figures below and in the next page provide full results of evaluation on Levenshtein distance, including precision, recall and F1-score at different thresholds, at threshold 0.4, and across different time periods in the CLEF-HIPE 2020 dataset. 

\begin{figure*}[h]
    \centering
    \includegraphics[width=\textwidth]{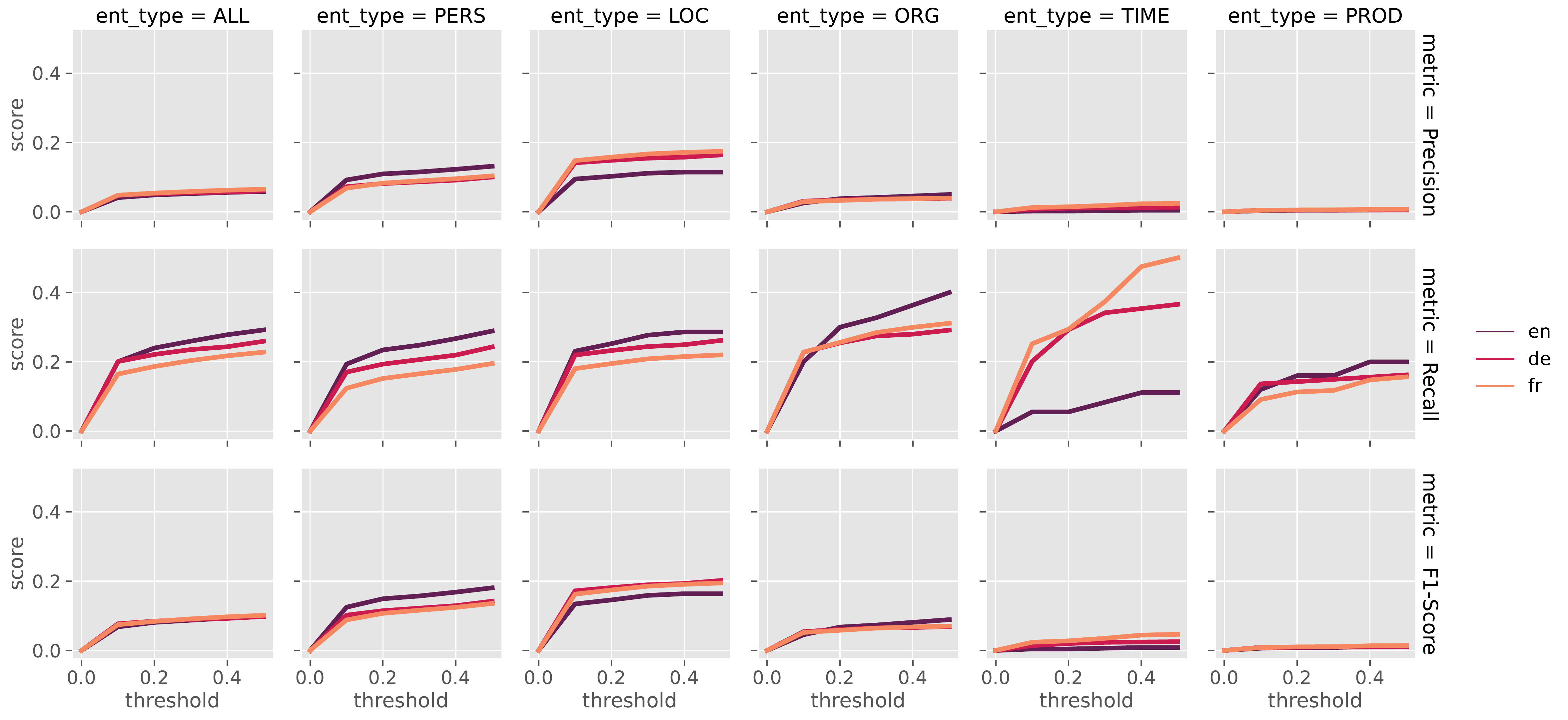}
    \caption{Precision, recall and F1-score (resp. first, second and third rows) at different Levenshtein distance thresholds and for different languages. Languages are distinguished by line color.}
    \label{fig:Lev_thresholds}
\end{figure*}

\begin{figure*}[h]
    \includegraphics[width=\textwidth]{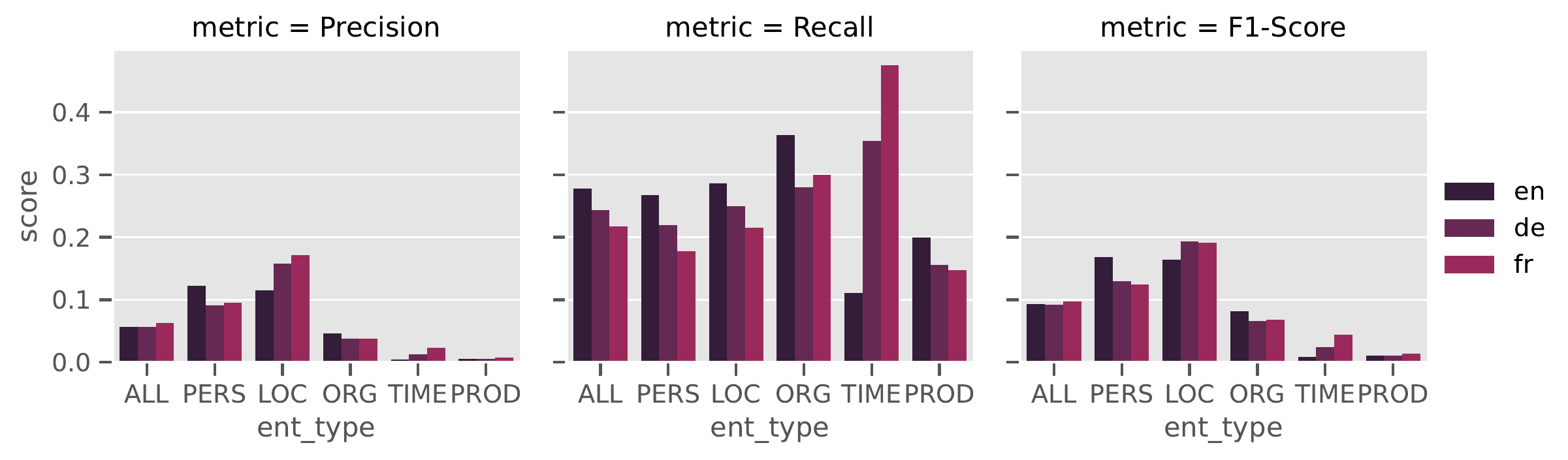}
    \caption{Precision, recall and F1-score (resp. first, second and third columns) by entity type at Levenshtein distance threshold 0.4 for different languages.}
    \label{fig:Lev_threshold_04}
\end{figure*}

\begin{figure*}[h]
    \centering
    \includegraphics[width=\textwidth]{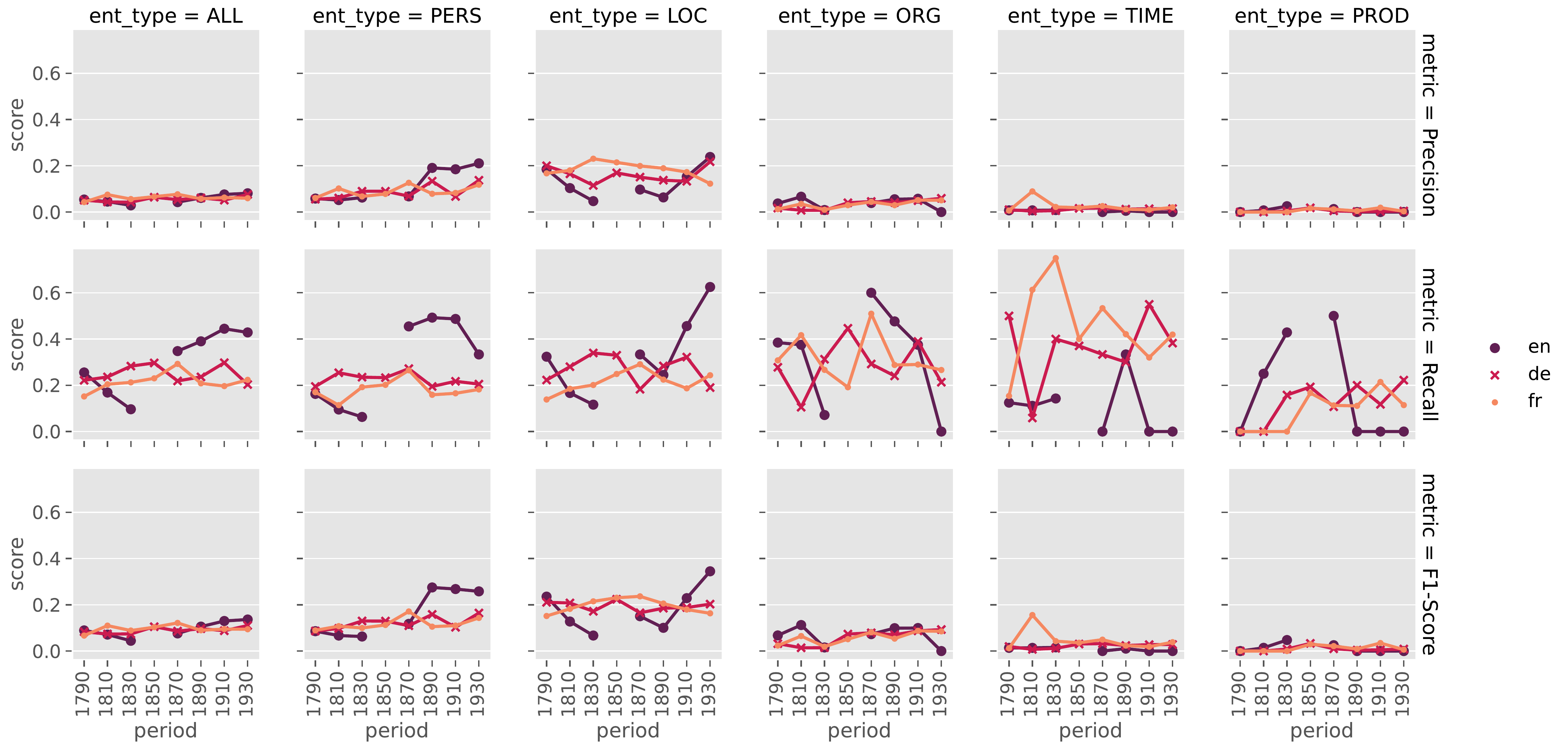}
    \caption{Precision, recall and F1-score (resp. first, second and third rows) at Levenshtein threshold 0.4 across periods for different languages. Languages are distinguished by both the line color and the type of dot.}
    \label{fig:Lev_threshold_04_period}
\end{figure*}

\end{document}